\documentclass[journal]{IEEEtran}

\usepackage{cite}
\usepackage{amsmath,amssymb,amsfonts}
\usepackage{algorithmic}
\usepackage{textcomp}
\usepackage{soul,color}
\usepackage{nicefrac}
\usepackage{subfigure}
\usepackage{multirow}
\usepackage{multicol}
\usepackage{tikz} 

\newcommand\submittedtext{%
  \footnotesize This work has been submitted to the IEEE for possible publication. Copyright may be transferred without notice, after which this version may no longer be accessible.}
\begin{document}
\title{



An End-to-End Deep Learning Generative Framework for Refinable Shape Matching and
Generation
}

\author{Soodeh Kalaie, Andy Bulpitt, Alejandro F. Frangi, and Ali Gooya
\thanks{S.Kalaie and A.Bulpitt are with the School of Computing, University of Leeds, Leeds, UK (e-mails: s.kalaie1@leeds.ac.uk; a.j.bulpitt@leeds.ac.uk), A.F. Frangi is with the Schools of Computer Science and Health Science at the University of Manchester, Manchester, UK (e-mail: alejandro.frangi@manchester.ac.uk),
A.Gooya is with the School of Computing Science, University of Glasgow, Glasgow, UK (e-mail: Ali.Gooya@glasgow.ac.uk)
}
}

\maketitle
\begin{tikzpicture}[remember picture,overlay]
    \node[anchor=south, text width=1.5\textwidth, align=center,yshift=10pt] at (current page.south) {\fbox{\parbox{\dimexpr0.65\textwidth-\fboxsep-\fboxrule\relax}{\submittedtext}}};
  \end{tikzpicture}%
\begin{abstract}
Generative modelling for shapes is a prerequisite for In-Silico Clinical Trials (ISCTs), which aim to cost-effectively validate medical device interventions using synthetic anatomical shapes, often represented as 3D surface meshes.
However, constructing AI models to generate shapes closely resembling the real mesh samples is challenging due to variable vertex counts, connectivities, and the lack of dense vertex-wise correspondences across the training data. Employing graph representations for meshes, we develop a novel unsupervised geometric deep-learning model to establish refinable shape correspondences in a latent space, construct a population-derived atlas and generate realistic synthetic shapes. We additionally extend our proposed base model to a joint shape generative-clustering multi-atlas framework to incorporate further variability and preserve more details in the generated shapes. Experimental results using liver and left-ventricular models demonstrate the approach's applicability to computational medicine, highlighting its suitability for ISCTs through a comparative analysis.

\end{abstract}


\section{Introduction}
\label{sec:introduction}

%
\IEEEPARstart{I}{n} recent years, the growing complexity and diversity of medical devices and technologies have posed significant challenges in evaluating and optimising their design and clinical applications. 
While clinical trials are ideal for such evaluations, they are not always practical due to various constraints, such as ethical limitations, costs, time requirements, and difficulties in recruiting enough subjects. 
 In-Silico Clinical Trials (ISCTs) are prime examples of applications that can benefit significantly from synthetic data and offer an alternative way to evaluate medical devices virtually \cite{abadi2020virtual}.
 
Machine learning methods, combined with computational modeling and simulation, provide valuable insights for assessing the safety and efficacy of new therapies and medical devices through ISCTs 
\cite{pappalardo2019silico}.
Virtual populations of anatomical shapes, typically represented as computational meshes/grid graphs, play a crucial role in facilitating ISCTs of clinical devices. Specifically, ISCTs require the generation of virtual patient populations that capture significant anatomical and physiological variability representative of the target patient populations effectively. This enables a meaningful in-silico assessment of device performance.

However, building rich or descriptive generative shape models from anatomical structures is challenging for multiple reasons.
The foremost challenge arises from the variability in real-world anatomical shapes derived from different subjects, leading to inconsistent mesh topologies and a lack of general topological correspondence. This poses a significant hurdle for existing techniques to generate coherent and anatomically plausible shape populations.
Moreover, existing techniques often require access to large volumes of training data containing the same semantic parts or shapes in each sample, necessitating expensive and laborious annotation of medical imaging data. 
Given that the morphology of organs across a population is highly heterogeneous, modelling and generating these shape variations in a virtual population is challenging. 
The increasing availability of large-scale medical imaging datasets has enabled these underlying shape variations in the population to be modelled more accurately. Due to the computational challenges of working with large sample sizes, traditional methods for shape analysis can be limited in practical application \cite{ng2014shape}. Hence, this study mainly focuses on the challenge of generating representative populations of anatomical shapes suitable for in-silico experiments.

In the field of shape processing, finding a meaningful shape correspondence is a fundamental task and is considered a difficult problem in numerous geometry processing applications, including shape generation \cite{van2011survey}.
Hence, more specifically, our aim is to address generative shape modelling (i.e. virtual population modelling) with the two following specific objectives:
\begin{itemize}
\item 
Addressing the issue of finding a dense correspondence map between the shapes with variable mesh typologies. We propose an unsupervised deep learning framework that learns correspondences from a population of spatially aligned training meshes with variable topology. 
\item 
Introducing a data-driven generative model for creating virtual populations of anatomical shapes that are diverse yet realistic. Our framework generates realistic synthetic shapes with variable anatomical topology, facilitating a more comprehensive exploration of anatomical variability in virtual populations.
\end{itemize}

\section{Related Work}
Conventional synthetic anatomical shape generation includes applying variants of Principal Component Analysis (PCA) on shape spaces within Statistical Shape Models (SSMs). In SSM training using PCA, point-based shape representations are commonly used due to their simplicity and independence from topology. Notable methods in this field include the active 
shape models and active appearance models by Cootes \textit {et al.}
\cite{cootes2001active}. 
While PCA-based statistical shape modeling has been widely applied to generate virtual anatomical populations \cite{young2009computational}, conduct quantitative shape analysis for computer-aided diagnosis \cite{shen2012detecting}, and facilitate model-based segmentation 
\cite{castro2015statistical}, 
there has been a growing interest in deep learning approaches for generative modelling in recent years. 

A few studies have employed these approaches to create virtual populations of anatomical structures \cite{danu2019deep,gutierrez2021discriminative,beetz2021generating,romero2021clinically}.
Danu \textit {et al.} \cite{danu2019deep} employed deep generative models (i.e. VAE and GAN)
for generating voxelised vessel surfaces, where they represented the unstructured surface mesh as a three-dimensional image. Therefore, it becomes inherently compatible with the standard convolutional neural network architecture. Although the results show potential on
employing deep neural network-based generative models on three-dimensional surfaces, they cannot deal with complex data structures.
In \cite{gutierrez2021discriminative}, a recent deep learning framework for anatomical shape analysis is presented, employing a conditional generative model. It operates on unordered point clouds with fixed cardinality, eliminating the need for point correspondences. While generating plausible shapes, evaluating fidelity poses challenges, focusing primarily on generalization without considering additional criteria for virtual population assessment.
Presenting cardiac biventricular anatomy as point clouds, Beetz \textit {et al.} \cite{beetz2021generating} introduced a geometric deep learning approach to generate populations of realistic biventricular anatomies. The personalisation of anatomies is further improved by incorporating subpopulation-specific characteristics as conditional inputs. It is important to note that these methods rely on point-cloud representations, which do not utilise the surface information embedded in meshes. In a different approach, Romero \textit {et al.} \cite{romero2021clinically} used binary masks of aortas as inputs to explore the generation efficiency of GANs. This approach explicitly generates a cohort of patients meeting a specific clinical criterion, even without access to a reference sample of that phenotype.

In summary, most of these approaches require laborious preprocessing to establish dense point correspondence among training shapes. Additionally, they are limited to datasets with identical topological structures. Therefore, achieving meaningful shape matching, or correspondence, within the training data is often necessary to utilize these techniques effectively.
Shape matching involves establishing meaningful point-wise correspondences between shapes. This task can be challenging due to the need for a comprehensive understanding of structures at both local and global scales. Due to its wide range of applications, 
a large body of research has been dedicated to addressing diverse variants of shape matching problems over the past decades \cite{van2011survey}.
Most techniques approach the problem as the registration of shapes as point sets, where correspondences are often derived from the proximity of aligned elements achieved through initial shape alignment. 
Research conducted in \cite{myronenko2010point,ma2020point} employs a probabilistic methodology to align two point sets using their Gaussian mixture representations.
These methods aim to determine spatial transformations across different shapes by minimising distances within these distributions. This approach involves a soft-assignment strategy, wherein correspondences are assigned probabilistically to points, encompassing a more comprehensive consideration of possible correspondences.
Different strategies are employed in \cite{rusinkiewicz2001efficient,zhang2021fast} to align point sets by optimizing local quadratic distances. However, due to the intricacy of deformations within the optimisation procedure, the optimisation can quickly get stuck in local optima. 

With recent advancements in deep learning, numerous methods for 3D shape matching have emerged. While some recent approaches \cite{eisenberger2023g,cao2022unsupervised} demonstrate accurate matchings for triangle meshes, they have not been applied to medical datasets. Additionally, these methods rely on specific mesh resolutions and canonical embeddings, potentially limiting generalisation to unseen test shapes \cite{cao2022unsupervised}.
More recent developments in graph matching have leveraged deep learning techniques to determine optimal point-to-point correspondences \cite{wang2019learning,zanfir2018deep}. 
These methods introduce supervised graph-matching networks, focusing on displacement rather than registration. It is worth noting that these proposed graph-matching techniques are applied offline and remain static during shape generation. In an alternative approach, Bai \textit{et al.} introduced a novel pair-wise graph matching framework in 
\cite{bai2019simgnn}, 
presenting a learning-based pair-wise graph matching framework as opposed to pairwise graph distance computations. 
However, their methods are constrained to computing similarity scores for complete graphs due to their reliance on spectral-based graph convolution techniques. To aggregate neighbouring information, they incorporated multi-scale graph convolutional layers and computed multiple similarity matrices, which escalates time complexity for handling extensive graph structures.

The present study is an extension of our preliminary work \cite{kalaie2023geometric} on virtual population modelling.
More specifically, here, we propose an Atlas Refinable Attention-based Shape Matching and Generation network (Atlas-R-ASMG), a framework that jointly learns high-quality refinable shape matching and generation using 3D surface mesh data while constructing a population-derived atlas model during the process.
This approach enables the generation of anatomical shapes that are both diverse and realistic, facilitating a more comprehensive exploration of anatomical variability in virtual populations.
This capability allows embedding extensive anatomical variability in virtual populations, 
holding the potential for 
more accurate and relevant in silico experiments and reliable insights in various medical and research fields.

Our contributions related to core methodologies include: 
\begin{itemize}
    \item Presenting an all-in-one hierarchical deep learning framework for atlas construction, shape matching and generation using geometric deep learning. The model learns to generate high-quality, plausible shapes from input training meshes with variable topologies. 
    
    \item The proposed shape matching is unsupervised and refinable, employing an attention mechanism to establish correspondences in latent space. This results in robust refinable vertex-wise correspondences and improvement over the baseline methods. 
    \item We further extend the Atlas-R-ASMG model to a joint shape clustering-generative framework (named mAtlas-R-ASMG) with improved generation performance and more comprehensive shape analysis potential via clustering.
    
 
\end{itemize}
While the proposed generative framework is demonstrated here on the Left Ventricle (LV) and liver structures, it is generic in design and can be applied similarly to other anatomical shape ensembles.

\section{Methods}
In this section, the base model (Atlas-R-ASMG) and its extension model (mAtlas-R-ASMG) are presented in detail.
\subsection{Generative Model: Atlas-R-ASMG}
\label{sec:Atlas-R-ASMMG}
\textbf{Problem Statement:} 
A shape given in the form of a mesh can be represented as a graph-like construct, denoted as $g = (V, \mathbf{A}, \mathbf{X})$, where $V$ is a finite set of nodes $V={1,2,...,N}$, and $\mathbf{A}$ is the adjacency matrix
$\mathbf{A} \in  \left \{0,1\right \} ^{N \times N}$.
Hence, in the remainder of this paper, we will use both shape and graph interchangeably. Suppose each vertex $v$ has a $d_x$-dimensional
feature vector $\mathbf{x}_{v}$, we use $\mathbf{X} \in \mathbb{R}^{N \times d_x}$ to denote the graph's node feature matrix with each row representing a vertex.
The corresponding stochastic latent variable of vertex $v$ is denoted as $\mathbf{z}_{v}\in \mathbb{R} ^{d_z}$, summarised in a matrix $\mathbf{Z} \in \mathbb{R} ^{N\times d_z}$.
\begin{figure*}[t]
    \centering
    \includegraphics[width = 1 \textwidth]{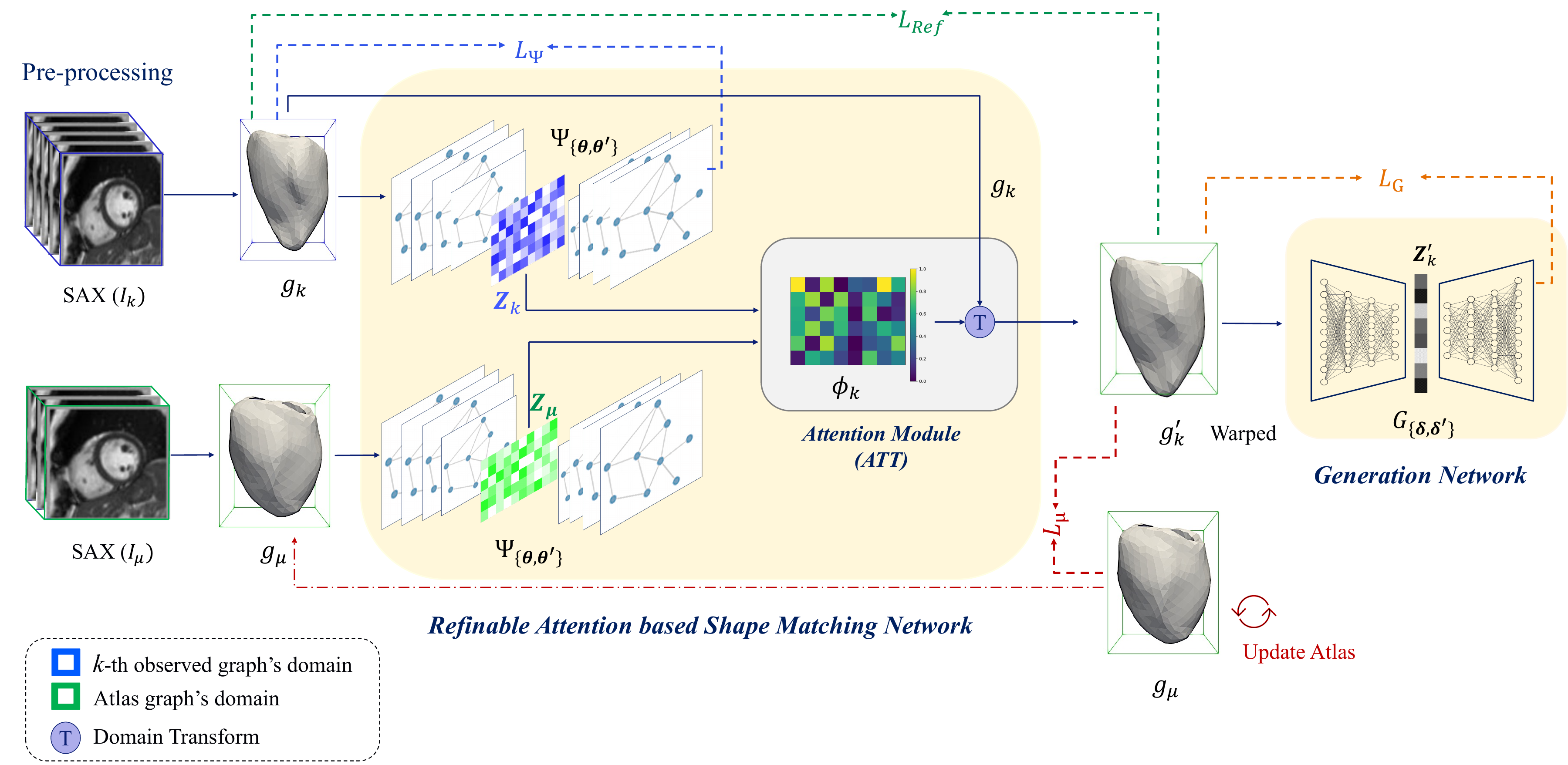} 
\caption{Diagram illustrating the overview of the Atlas Refinable Attention-based Shape Matching and Generation network (Atlas-R-ASMG) framework, designed for shape generation from shapes with variable mesh topology. Pre-processing involves extracting input shapes of different sizes from 3D images.}
    \label{fig:diagchap5}
\end{figure*}
Consider a training graph dataset ${G} = \{ g_k\}_{k=1}^{K}$ containing $K$ shapes with variable topologies/cardinalities. Each $g_k$ represents a 3D surface shape with cardinality $|V_k|=N_k$.
Our goal is to train a learning framework that promotes two critical characteristics of the generated shapes: \textit{high-quality shape matching} and \textit{anatomical plausibility}. 
In the rest of the paper, we indicate the atlas graph as $g_{\mu}$, which is computed within the proposed framework and facilitates vertex-wise correspondences. This population-derived atlas serves as a comprehensive reference template, encapsulating the common features and variations observed within the dataset.

As illustrated in Fig. \ref{fig:diagchap5}, the Atlas-R-ASMG pipeline comprises two primary components that are jointly trained in an end-to-end manner. 
The initial component of the pipeline involves establishing correspondence maps across the training data using an unsupervised probabilistic refinable attention-based shape-matching network. 
This process simultaneously refines the atlas shape and warps it into each observed shape. We refer to this as the domain transformation (i.e., normalisation of shape structures) within the 3D space. 
This normalisation procedure ensures a consistent representation of shapes across different instances of input meshes, accommodating topological/structural variations and facilitating effective shape matching.
The second component involves a generative network that learns a probability density function based on a set of structurally normalised shapes. This generative network captures the underlying distribution of the normalised shapes in the 3D space, enabling the generation of new shapes that adhere to the learned distribution.

The Atlas-R-ASMG pipeline effectively learns to fit a population-derived atlas shape to the training data through the end-to-end training process. Integrating the refinement procedure and jointly inferring the population-derived atlas contributes to improved shape matching, generation, and overall performance. 
We present the first component of the pipeline, Refinable Attention-based Shape Matching (R-ASM) in Section \ref{sec:R-ASMG}. This is followed by integration of the second component, the Generation Network, in Section \ref{sec:generation}. We next introduce the Unsupervised Loss in Section \ref{sec:unsuploss}, and the method for constructing the atlas
in Section \ref{sec:atlas}.


\subsubsection{Refinable Attention-based Shape Matching (R-ASM)}
\label{sec:R-ASMG}
This section focuses on developing a  learning-based shape-matching approach that employs graph neural networks and attention mechanisms. The matching procedure aims to establish vertex-wise correspondences (in latent space) by measuring the similarity of local node embeddings, eliminating the need to solve an optimisation problem during inference.

\subsubsection*{Feature extraction network}

A variational graph autoencoder model with spatial-based graph convolution layers is used to build the feature extraction network $\Psi$ (parametrised by trainable encoder/decoder weights of $\{\boldsymbol{\theta},\boldsymbol{\theta}^{'}\}$) and learn nodal embeddings.
Taking inspiration from \cite{verma2018feastnet}, our convolution layers employ a spatial-based graph convolution operator that, unlike previous work \cite{monti2017geometric},
does not rely on static pre-defined local pseudo-coordinate systems over the graph but instead dynamically determines the correspondences between filter weights and graph neighbourhoods.
This convolution operator utilises soft correspondence for each neighbour of a node $v$, rather than assigning each neighbour to a single weight matrix.

This network takes the adjacency matrix $\mathbf{A}$ and node features $\mathbf{X}$ as input and tries to recover the feature matrix $\mathbf{X}$ through the hidden layer embeddings $\mathbf{Z}$.
Given the template/atlas $g_{\mu}$ and observed
shapes $g_{k}$, nodal embeddings $\mathbf{Z}_{\mu}$ and $\mathbf{Z}_k $ are computed by a shared graph convolutional neural network (GCN) $\Psi$ as  
\begin{equation}
    \{\mathbf{Z}_k,\mathbf{Z}_{\mu}\}= \Psi_{\{\theta,\theta^{'}\}}(g_k,g_{\mu})
     \label{eq:FE} 
\end{equation}
where $\mathbf{Z}_k= \Psi_{\theta}(\mathbf{X}_k,\mathbf{A}_k)$ and $ \mathbf{Z}_{\mu}= \Psi_{\theta}(\mathbf{X}_{\mu},\mathbf{A}_{\mu})$.
The feature extraction network $\Psi$ can be presented in two different settings: sGCN and hGCN.
The former refers to where representations of vertices are only defined by spatial features (i.e. positions of vertices), while the latter serves as hybrid representations of vertices defined by spatial positions of vertices along with the associated vertex normal vectors.

\subsubsection*{Attention module}
Once the graph embedding is learned, an attention mechanism is applied to those embedded features to learn correspondences across the population. The ATT module learns the mapping/correspondence map $\mathbf{\phi}_k$ to warp each observed $k$-th shape to the atlas and obtain a normalised and structurally consistent shape as 
\begin{equation} g'_k=ATT(\underbrace{\mathbf{Z}_k,\mathbf{Z}_{\mu}}_\text{$\mathbf{\phi}_{k}$},g_k);
 \label{eq:ATT}   
\end{equation}
where $g_k \in \mathcal{F}(\mathbb{R}^{N_k})$ and $g'_k \in \mathcal{F}(\mathbb{R}^{N_{\mu}})$.
This process is referred to as domain transformation from the node function space $\mathcal{F}(\mathbb{R}^{N_k})$ to $\mathcal{F}(\mathbb{R}^{N_{\mu}})$, which results in normalising the structure of the input shape. Each normalised shape $g'_k$ is represented by a matrix feature $\mathbf{X}^{'}_{k}$. 
Given latent node embeddings $\mathbf{Z}_{k} \in \mathbb{R}^{N_k \times d_z}$ and $\mathbf{Z}_{\mu} \in \mathbb{R}^{N_{\mu} \times d_z}$, we obtain soft correspondences (attention maps) $\mathbf{\phi}_{k} \in [0,1]^{N_{\mu} \times N_{k}}$ using the mapping function $\mathit{\phi}_{k} = \mathit{Softmax}(\lambda \mathbf{Z}_{\mu}\mathbf{Z}_{k}^{T})$ in the embedded-space paradigm. The module employs a soft attention mechanism as a shape's domain transformer, normalising shape structures using $\mathbf{{x}'}_{kj} = \sum_{i=1}^{|V_k|} {\mathit{\phi}}_{kji}\mathbf{x}_{ki} \; ; j=\{1,...,|V_\mu|\}$.



A refinement procedure is integrated into the pipeline during the training phase to ensure topological plausibility and accuracy. This iterative refinement step addresses errors in vertex-wise correspondences and shape normalisations, progressively improving the alignment and matching of shapes. The refinement strategy plays a crucial role in enhancing the overall performance and accuracy of the pipeline.

\subsubsection{Generation Network}
\label{sec:generation}
Next, a variational autoencoder denoted by $G$ (parameterised by a a set of encoder/decoder parameters $\{\boldsymbol{\delta},\boldsymbol{\delta}^{'}\}$) is trained using the structure normalised shapes to generate a set of synthetic shapes $g^{syn}$ as;
\begin{equation}
   \{ g^{syn}\}=G_{\{\boldsymbol{\delta},\boldsymbol{\delta}^{'} \}}(g'_k).
 \label{eq:G}   
\end{equation}
This network employs fully-connected layers in a $\beta-$VAE structure. To generate novel shape samples, we sample from the prior normal distribution and feed this sample into the decoder $G_{\{\boldsymbol{\delta}'\}}$, which produces a random synthetic shape having the same connectivity as the atlas.

A forward pass through the network via Feature extraction network \eqref{eq:FE}, Attention module
\eqref{eq:ATT}  and Generation network \eqref{eq:G} can be
expressed concisely as:
\begin{equation}
   \{{g}^{syn}\}=G_{\{\boldsymbol{\delta},\boldsymbol{\delta}^{'} \}}\Bigg(ATT \bigg(\Psi_{\{\boldsymbol{\theta},\boldsymbol{\theta}^{'}\}}(g_k,g_{\mu}),g_k \bigg) \Bigg).
 \label{eq:forwardpass}   
\end{equation}

\subsubsection{Unsupervised Loss}
\label{sec:unsuploss}


The overall unsupervised loss is composed of three parts: 
feature extraction loss ($L_{\Psi}$), refinement loss ($L_{Ref}$), and generation loss ($L_G$), i.e., 
$L = L_{\Psi}+L_{Ref}+ L_G$.
Loss terms $L_{\Psi}$ and $L_{Ref}$ are associated with shape matching procedure and $L_G$ refers to generation process (see Fig. \ref{fig:diagchap5}). In the following, each loss term is explained in more detail.

\textbf{Feature extraction loss}: 
In the variational graph autoencoder network $\Psi$, we parameterise the approximate posterior $q_{\boldsymbol{\theta}}(\textbf{Z}|\textbf{X},\textbf{A})$ with an encoder and the likelihood $p_{\boldsymbol{\theta}^{'}}(\textbf{X}|\textbf{Z},\textbf{A})$ with a decoder.
This network is trained in an unsupervised manner by optimising the variational lower bound $L_{\Psi}$ w.r.t. the variational parameters:
\begin{align}
\label{eq:lossPsi}
L_{\Psi}(\boldsymbol{\theta},\boldsymbol{\theta}^{'} ) & =  \sum_{k=1}^{K}
\big( \mathbb{E}_{q_{\boldsymbol{\theta}}(\textbf{Z}_k|\textbf{X}_k,\textbf{A}_k)} [\log p_{\boldsymbol{\theta}^{'}}(\textbf{X}_k|\textbf{Z}_k,\textbf{A}_k)] \notag \\
& \hspace{2mm}- w_{kl}D_{KL} [ q_{\boldsymbol{\theta}}(\textbf{Z}_k|\textbf{X}_k, \textbf{A}_k)\parallel p(\textbf{Z}_k)]\big);
\end{align}
where the first term quantifies reconstruction accuracy, evaluating how well the decoder network $p_{\boldsymbol{\theta}^{'}}(\textbf{X}_k|\textbf{Z}_k, \textbf{A})$ reconstructs shape $\mathbf{X}_k$ from latent space values $\mathbf{Z}_k$.
The second term, the regularisation term, tends to regularise the organisation of the latent space. $D_{KL}$ is the Kullback-Leibler divergence (KL divergence) between the approximate posterior $q_{\boldsymbol{\theta}}(\textbf{Z}_k|\textbf{X}_k, \textbf{A}_k)$ and the prior distribution $ p(\textbf{Z}_k)$, weighted by $w_{kl}$. 
A unit Gaussian distribution defines the prior distribution $p(\textbf{Z}_k)=\prod_{n=1}^{N_k} \textit{N}(\textbf{z}_{kn};\textbf{0},\textbf{I})$.


In the two settings (s/h)GCN; the network $\Psi$ (named sGCN) learns embeddings of the spatial information and node feature matrix $\mathbf{X}$ denotes the cartesian geometry of nodes, whose $i$-th row is $\mathbf{x}_{i} = (x_{i},y_{i},z_{i})$.
The network $\Psi$ (named hGCN) learns combined embeddings of geometry and surface. The feature matrix $\mathbf{X}$ is a horizontal concatenation of vertex
geometries and vertex normals and $i$-th row represented by $\mathbf{x}_{i} = (x_{i},y_{i},z_{i}, \mathbf{n}_{i}^x,\mathbf{n}_{i}^y,\mathbf{n}_{i}^z)$ where $\mathbf{n}$ denotes normal vector \cite{thurrner1998computing}. In this setting, the regularisation term consists of $D_{KL}$ and 
the normal consistency term $w_{norm} L_{norm} (g_k^{rec})$, for computing the normal consistency between neighboring faces in $g_k^{rec}$ reconstructed mesh \cite{pytorch3d}. Normal consistency measures how smoothly the normals of adjacent faces align with each other along shared edges.
Hyperparameters $w_{kl}$ and $w_{norm}$ are empirically set to $1e^{-3}$ and $1e^{-2}$, respectively, for both datasets.
\textbf{Refinement loss}: 
Shape matching incorporates a refinement strategy to avoid finding false correspondences while projecting shapes onto the atlas domain. To achieve this, the loss $L_{Ref}$ encourages accurate transformation of shape $g_k$ to the atlas domain and is defined as 
\begin{equation}
\label{eq:lossRef}
   L_{Ref}=\sum_{k=1}^{K} w_{cd}CD(g_k,g'_k) +w_{lap}L_{lap}(g'_k);
\end{equation}
where Chamfer Distance (CD) measures the distance of vertices between two graphs: $CD(g,g')=\sum_{\mathbf{x}_{i} \in g} \text{min}_{{g}'} 
| \mathbf{x}_{i}- \mathbf{x}'_{j}|^2 +\sum_{\mathbf{x}'_{j}\in {g}'} \text{min}_{g} | \mathbf{x}_{i}- \mathbf{x}'_{j} |^2 $.
The Laplacian loss $L_{lap}$ is a regularization term that encourages neighbouring vertices to move coherently, reducing mesh self-intersections and allowing for smoother surface reconstructions\cite{nealen2006laplacian}. 
This loss measures the difference
between the vertex position and the mean over all neighbouring vertices. 
For the LV dataset, the hyperparameters $w_{cd}$ and $w_{lap}$ are empirically set to $1$. For the liver dataset,
$w_{cd}$ is set to 1, and $w_{lap}$ is set to 1.2 through empirical tuning.
As a result, the parameters of the shape matching network are optimised to minimise both losses $L_{\Psi}$ and $L_{Ref}$.

\textbf{Generation loss}: To generate synthetic shapes, the loss function ${L}_{G}$ follows the original loss of the $\beta$-VAE, where hyperparameter $\beta$ makes a balance between low reconstruction error and high latent space quality, which emphasises discovering disentangled latent factors.
This generative network derives a pdf from the set of normalised graphs. The probability of node variations is approximated in the latent space $\textbf{Z}_{k}^{'}\in \mathbb{R}^{L}$ via a posterior $q_{\boldsymbol{\delta}}(\textbf{Z}_{k}^{'}|\mathbf{X}_{k}^{'})$. By drawing samples from this approximate posterior probability, we estimate the likelihood of the observed population. With regard to the variational parameters $\boldsymbol{\delta},\boldsymbol{\delta}^{'}$, we optimise the variational lower bound
\begin{align}
\label{lossVAE}
L_G(\boldsymbol{\delta},\boldsymbol{\delta}^{'}) & = \sum_{k=1}^{K}{( \mathbb{E}_{q_{\boldsymbol{\delta}}(\textbf{Z}_{k}^{'}|\mathbf{X}_{k}^{'})} [\log p_{\boldsymbol{\delta}^{'}}(\mathbf{X}_{k}^{'}|\textbf{Z}_{k}^{'})] } \notag \\
&- \beta D_{KL} [ q_{\boldsymbol{\delta}}(\textbf{Z}_{k}^{'}|\mathbf{X}_{k}^{'})\parallel p(\textbf{Z}_{k}^{'})]);
\end{align}
where $\mathbf{X}_{k}^{'}\in \mathbb{R} ^{N_{\mu}\times 3} $ presents graph geometric features after the normalisation process, and the first term is a reconstruction error, which computes the squared Euclidean distance between input and reconstructed shapes by the decoder.
$D_{KL}$ denotes the Kullback-Leibler and computes the divergence between the Gaussian prior ${N}(0, I)$ and posterior distributions of the latent space $\mathbf{Z}'$
and the KL divergence is weighted by $\beta$. 
The optimal value for $\beta$ was achieved by empirically decreasing the values from $2e^{-3}$ to $2e^{-6}$ for the LV dataset and from $1e^{-3}$ to $2e^{-3}$ for the liver dataset.




\subsubsection{Atlas construction}
\label{sec:atlas}

It is important to note that the shape matching and the generation are trained jointly. The normalised shapes $g'_k$, for all $k$, and the atlas shape $g_{\mu}$ are updated at the end of each epoch by warping the training shapes to atlas space via a forward pass of Atlas-R-ASMG and averaging across the samples.
The atlas (represented by the feature matrix $\mathbf{X}_{\mu}$) is reconstructed by minimising 
the following cost function, which entails a Laplacian term for smoothness 
\begin{align}
	\label{eq:lossmu}
	 L_{\mu}  = \frac{1}{2}\sum_{k=1}^{K}\sum_{j=1}^{N_{\mu}} |\mathbf{x'}_{kj}-\mathbf{x}_{\mu j} |^{2}  +  \frac{\gamma}{2}\sum_{j=1}^{N_{\mu}}\sum_{q=1}^{N_{\mu}} a_{jq} |\mathbf{x}_{\mu j}-\mathbf{x}_{\mu q}|^{2};
\end{align}
where $a_{jq}$ are elements of the adjacency matrix $\mathbf{A}_{\mu}$. 
In order to minimise $L_\mu$, starting from an initial canonical atlas shape $\mathbf{X}_{\mu}^{(0)}$, a new atlas $\mathbf{X}_{\mu}^{(i+1)}$ is iteratively computed from $\mathbf{X}_{\mu}^{(i)}$ according to 
\begin{align}
	\label{eq:atlas} 
	\mathbf{x}_{\mu j}^{(i+1)} \leftarrow \frac{\sum_{k}^{}\mathbf{x^{'}}_{kj}+\gamma\sum_{q\in\mathcal{N}_{\mu j}}^{}\mathbf{x}_{\mu q}^{(i)}}{K+\gamma\sum_{q\in\mathcal{N}_{\mu j}}^{}a_{jq} }; \qquad \forall j \in N_{\mu} 
\end{align}
where $\mathcal{N}_{kj}$ is the set of neighbours of $j$-th vertex and $\gamma=\nicefrac{N_{\mu}}{\max{(\mathcal{N}_{\mu j}})}$. We build the atlas shape $g_{\mu}$ with a fixed adjacency matrix $\mathbf{A}_{\mu}$ so that the vertex positions are optimised while preserving the topology of the atlas.

At the end of each training epoch, a forward pass through the
network \eqref{eq:forwardpass} is used to warp each training case to the atlas space.
The updated atlas shape at the end of each epoch, $(i >0)$ in
\eqref{eq:atlas}, is utilised during the following epoch for computing the losses $L_{\Psi}$ \eqref{eq:lossPsi} and $L_{Ref}$ \eqref{eq:lossRef}. This procedure results in both refined atlas, vertex correspondences, and eventually generated shapes.


The proposed Atlas-R-ASMG model considers a unique atlas for each shape category, warping all shapes onto a common atlas for correspondence map computation, resulting in meshes with the same connectivity (i.e. same mesh topology). However, in cases of large variability in appearance and anatomical structure within the shape population, a single representative atlas may not be suitable, leading to increased distance errors and diminished fidelity.
\subsection{Joint-Clustering Generative Model: mAtlas-R-ASMG}
\label{sec:mAtlas}
To address the above limitations, this section presents an extension of the model in Section \ref{sec:Atlas-R-ASMMG}; named Multi-Atlas Refinable Attention-based Shape Matching Generative model (mAtlas-R-ASMG). A "joint-clustering generative" shape model, along with multi-atlas construction for a specific organ, allowing for wider structural variations. Incorporating the "multi-atlas" shape matching procedure improves correspondence establishment and enables grouping shapes into meaningful clusters based on similarity measures, facilitating analysis of variation and complex spatial relationships. Additionally, incorporating multi-atlas construction with variable mesh topologies enhances the generative model's ability to synthesise virtual shape populations with diverse topological variations.
Therefore, the generative model can generate more morphological variability of shapes due to the presenting different shape clusters.
This is particularly useful in cases where vertex-to-vertex correspondences are challenging or inaccurate due to occlusions or variations in point density.
Overall, this model can (1) cluster shapes by capturing similarity between them, (2) preserve small details of the input shapes by assigning the best atlas (i.e. best mesh topology) to each observed shape, and (3) generate a multi-resolution virtual population of organ anatomies (i.e. variable topology in anatomical structures).

\subsubsection{Clustering Scheme}
\label{sec:clustering}
%

Assuming an unlabeled shape set ${G} = \{ g_k\}_{k=1}^{K}$ is projected into $M$ clusters. The shape set $\{g_{\mu_m}\}_{m=1}^M$ illustrates atlases (i.e. the average shapes) in each cluster with matrix representation $\mathbf{X}_{\mu_m} \in \mathbb{R}^{{N}_{\mu_m} \times 3}$. 

Projecting each shape on different atlases, the framework presents $M$ clusters of structurally normalised shape populations where each population presents $K$ shapes in the resolution of $|V|=N_{\mu_m}$. 
The shape $g_{k}^{'m}$ denotes normalised shape $k$ in the cluster $m$ and $\textrm{w}_{km}$ is its associated projection weight.
The weights elements $\textrm{w}_{km}$ are defined as follows:
\begin{align}
	\label{eq:cluster-weight} 
	\textrm{w}_{km}=\frac{e^{-\alpha d_{k}^{m}}}{\sum_{m=1}^{M}{e^{-\alpha d_{k}^{m}}}} ;
\end{align}
where $d_{k}^{m}=\frac{1}{N_{\mu_m}}\sum_j|\mathbf{x}_{kj}^{'m}-\mathbf{x}_{\mu_mj}|^2$ presents the mean squared error between each normalised shape and mean shape in each cluster and the normalised shape $g_{k}^{'m}$ is presented by the geometric feature matrix $\mathbf{X}_{k}^{'m}$. The $\alpha$ is a hyperparameter that controls the de-weighting degree in terms of dissimilarity. 
Therefore, a higher weight is assigned to a shape that is more similar to the $m$-th atlas. 
Shape-wise weights are denoted as vector $\textrm{w}_k \in \mathbb{R}^{M}$, and weight matrix $\textrm{W}=\left(\textrm{w}_{1}, \textrm{w}_{2}, ...,\textrm{w}_{K}\right)^T$ fulfills the constraint $\sum_{m \in M} \textrm{w}_{km} =1;  \forall k \in K$. 
For experimental settings, we found $\alpha =\frac{1}{std(\textrm{d})}$ is experimentally appropriate, where the distance matrix $\textrm{d}=\left({d}_{1}, {d}_{2}, ...,{d}_{K}\right)^T$ and $d_k \in \mathbb{R}^{M}$.

Similar to \eqref{eq:lossmu}, atlas shapes are reconstructed by minimising the following cost function 
\begin{align}
\notag
	\label{eq:lossmu-multi}
	 	 L_{\mu} & = \frac{1}{2}\sum_{m=1}^{M}\sum_{k=1}^{K}\sum_{j=1}^{N_{\mu_m}} \textrm{w}_{km}|{\mathbf{x}_{kj}^{'m}}-\mathbf{x}_{\mu_m j} |^{2} \\
  &+  \frac{\gamma_m}{2}\sum_{j=1}^{N_{\mu_m}}\sum_{q=1}^{N_{\mu_m}} {a_{\mu_m}}_{jq} |\mathbf{x}_{\mu_m j}-\mathbf{x}_{\mu_m q}|^{2};
\end{align}
where ${a_{\mu_m}}_{jq}$ are elements of the adjacency matrix $\mathbf{A}_{\mu_m}$ (demonstrates the topological structure of  cluster $m$).
For $\forall m \in M$, a new atlas $\mathbf{X}_{\mu_m}^{(i+1)}$ is iteratively computed from $\mathbf{X}_{\mu_m}^{(i)}$ according to
\begin{align}
	\label{eq:multiatlas} 
	\mathbf{x}_{\mu_m j}^{(i+1)} \leftarrow \frac{\sum_{k}^{}\textrm{w}_{km}{\mathbf{x}_{k}^{'m}}_j+\gamma_m\sum_{q\in\mathcal{N}_{\mu_m j}}^{}\mathbf{x}_{\mu_m q}^{(i)}}{\sum_{k}^{}\textrm{w}_{km}+\gamma_m\sum_{q\in\mathcal{N}_{\mu_m j}}^{}{a_{\mu_m}}_{jq} }; \forall j \in N_{\mu_m} 
\end{align}
where $\mathbf{x}_{\mu_m j}$ is a feature vector corresponding to $j$-th node in the $m$-th atlas graph and 
$\gamma_m=\nicefrac{N_{\mu_m}}{\max{(\mathcal{N}_{\mu_m j}})}$. We build the atlas shape $g_{\mu_m}$ with a fixed adjacency matrix $\mathbf{A}_{\mu_m}$.
Initialising atlases relies on canonical shapes guided by clinical experts' insights, enhancing the model's ability to generate clinically relevant clusters aligned with anatomical variations. This incorporation of expert knowledge ensures a robust and clinically meaningful multi-clustered learning process. 

Following \eqref{eq:multiatlas}, the population can generate $M$ atlas shapes with diverse topologies/cardinalities, showcasing variations in shapes. This approach addresses the limitation of relying on a single template shape for normalising real shapes, which may overlook certain details. By deriving multiple atlas shapes, the method captures a broader range of shape variations, enhancing the accuracy and comprehensiveness of the analysis.
\begin{table*}[!t]
\renewcommand{\arraystretch}{1.3}
\setlength{\tabcolsep}{10pt}

\caption{Shape Matching Quality: comparison between different methods; RSM, ASM and R-ASM using two distance metrics $HD$ and $CD$ (mean $\pm$ std) in $[mm]$. \textbf{Bold} numbers are the best and the second best with the best also underlined, for a given metric. Statistically significant (p-value $<0.01$ ) improvement of the best  model over a given model is indicated by superscript $\ast$.}
\centering
\begin{tabular}{ll ccc cc }
\hline \hline
&   &   &  \multicolumn{2}{c}{ASM}  &  \multicolumn{2}{c}{R-ASM}      \\ \cline{4-7}
 \text{Data} & \text{Metric}& \text{RSM} & \text{sGCN-ATT} & \text{hGCN-ATT} & \text{sGCN-ATT} & \text{hGCN-ATT}\\ \hline
 \multicolumn{1}{l}{\multirow{2}{*}{\text{LV}}}&$HD$& ${8.11 \pm 2.13}$ & ${8.32 \pm 1.77} $&   ${6.43 \pm 1.44}^* $ & $\mathbf{6.55+1.56}$ & $  \underline{\mathbf{5.91+1.38}}^*$\\ 
 & $CD$ & ${12.04 \pm 27.63}$& ${9.91 \pm 1.54}$ & ${9.85 \pm 0.94}^*$  & $\underline{\mathbf{4.06+0.56}}^*$ & $  \mathbf{5.01+0.40}$  \\ \hline 
  \multicolumn{1}{l}{\multirow{2}{*}{\text{Liver}}}&$HD$& ${35.09 \pm 14.55}$ & ${32.79 \pm 10.16} $&   ${27.30 \pm 7.71}^* $ & $ \mathbf{23.97 \pm 8.58} $ &$\underline{\mathbf{20.21 \pm6.53}}^*$ \\ 
 & $CD$ & ${254.46 \pm 211.57}$& ${139.66 \pm 82.66}$ & ${94.75 \pm 35.99}^*$ & $\mathbf{70.92 \pm 37.40}$ & $\underline{\mathbf{58.14 \pm 21.96}}^*$  \\ \hline \hline
\end{tabular}
\label{table:Rec-chap5}
\end{table*}
\begin{figure*}[t]
\centering     
\subfigure[sGCN-ATT]{\label{fig:scatter-sGCN}\includegraphics[width = 0.5 \textwidth ]{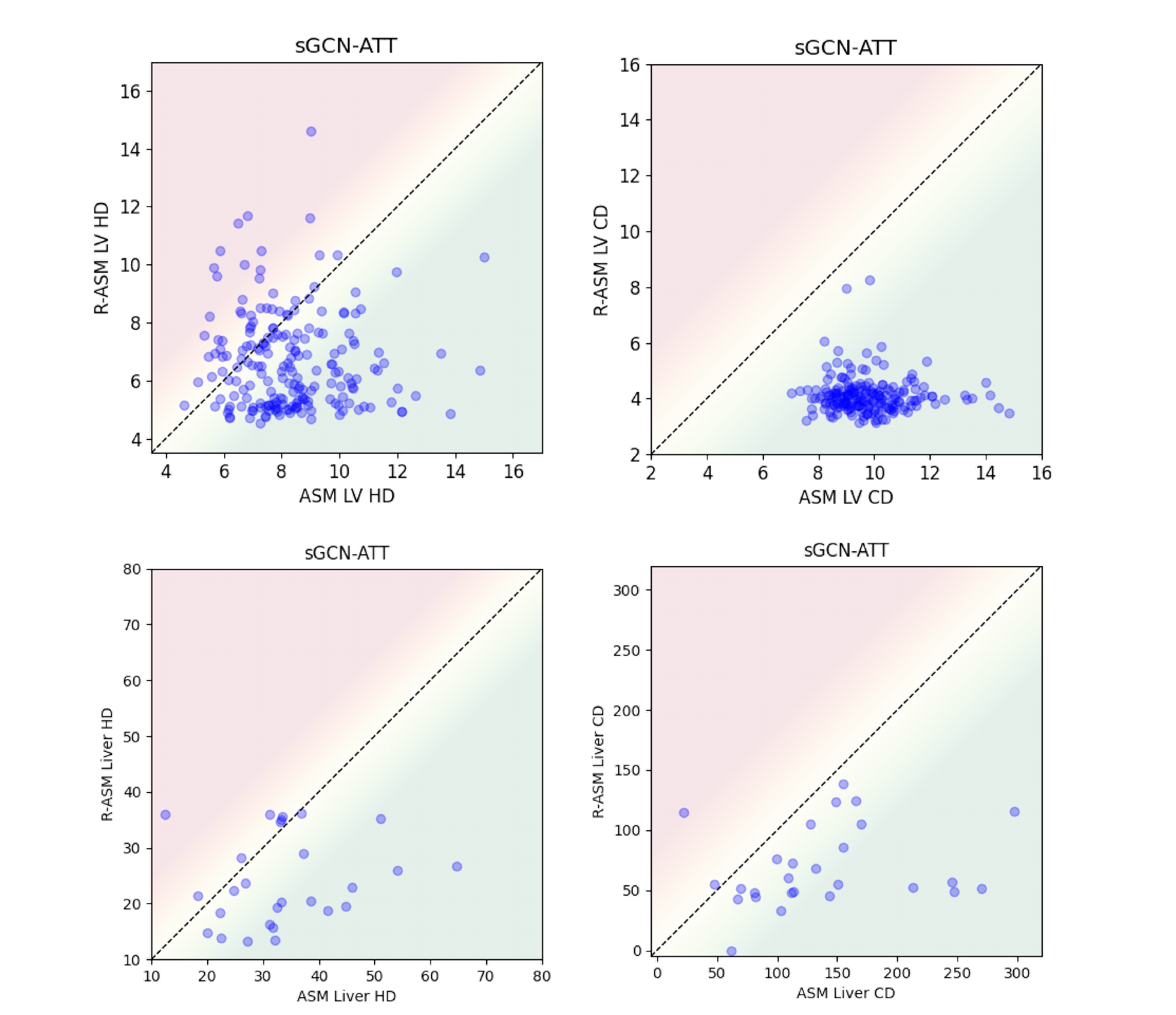}}  
\hspace{-8mm}
\subfigure[hGCN-ATT]{\label{fig:scatter-hGCN}\includegraphics[width = 0.5 \textwidth ]{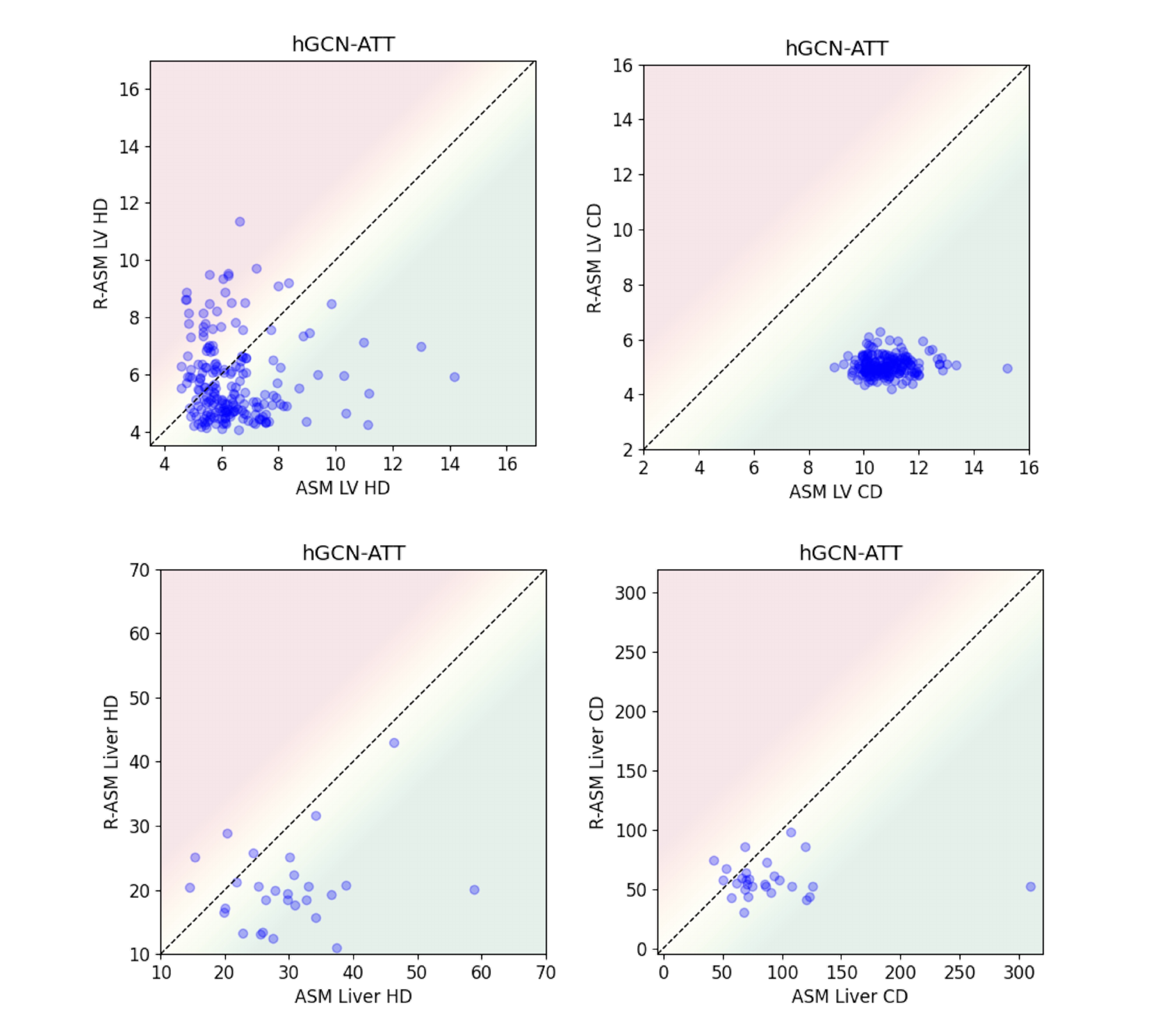} }\\
\caption{Scatter plots showing $HD$ (left) and $CD$ (right)  results of attention-based shape matching frameworks in $[mm]$, with (a) sGCN-ATT and (b) hGCN-ATT settings, on the $200$ LV and $28$ liver cases test set, comparing the ASM ($x-$axis) versus refinable model R-ASM ($y-$axis). The green/red gradients indicate an increase/decrease in performance with refinement, respectively. 
    Refinement generally improves $HD$ and $CD$ errors of attention-based shape-matching results. Degradation is observed for some LV/liver cases in terms of $HD$, whereas improvements in $HD$ can be significant.
}
\label{fig:synthticshapes}
\end{figure*}
\subsubsection{Generative Modeling}
Here, we train a multi-generative model given different clusters of shape populations to generate virtual populations with variable topology in anatomical structures.
More specifically, for $M$ clusters, our framework presents $M$ generative models where each generative network (designed in $\beta$-VAE structure) is trained to derive a pdf from a set of normalised graphs. Therefore, similar to the Section \ref{sec:generation}, for $m$-th generative model,
we optimise the variational lower bound $L_{G_m}$ w.r.t the variational parameters $\boldsymbol{\delta_m},\boldsymbol{\delta}^{'}_m$:
\begin{align}
\notag
    L_{G_m} &=\frac{1}{2}\sum_{k=1}^{K}\textrm{w}_{km}(\sum_{j=1}^{N_{\mu_m}} \left | {\mathbf{x}_{kj}^{'m}}^{rec}-{\mathbf{x}_{kj}^{'m}} \right |^{2} \\
    & - \beta_m D_{KL} [ q_{\boldsymbol{\delta}_m}(\mathbf{Z}_{k}^{'m}|\mathbf{X}_{k}^{'m})\parallel p(\mathbf{Z}_{k}^{'m})]);
\end{align}
where ${\mathbf{X}_{k}^{'m}}$ and ${\mathbf{X}_{k}^{'m}}^{rec}$ denote the input shape and its reconstruction respectively.
Considering equation \eqref{eq:cluster-weight}, the shape that is more relevant to cluster 
$m$ is assigned a higher weight. As a result, it will exert more influence on the $m$-th generative model compared to other generative models.
Finally, to generate a virtual population of shapes with variable topology in anatomical structures, our objective is to extract a sample from the joint distribution $p(\mathbf{Z}',m)$ where $m$ is the cluster value. Leveraging the dependencies among the variables, we employ ancestral sampling \cite{bishop2006pattern} to generate 3D surface mesh samples from our generative model. This joint distribution can be written as $p(\mathbf{Z}'|m)p(m)$ where $p(m)$ is a categorical distribution of M clusters, $p(m)=\frac{\textrm{w}_{m}}{\sum_{m}\textrm{w}_{m}}$, $p(\mathbf{Z}'|m)$ modelled by the decoder networks, allowing us to sample from models of $M$ clusters with latent dimension $L$.
\section{Datasets}
The methods are applied to two clinical datasets: 1000 LV shapes extracted from cardiac magnetic resonance (CMR) images of the UK Biobank 
\cite{petersen2015uk} 
(access application number 11350) and 139 liver shapes from the public CT-ORG image dataset from The Cancer Imaging Archive (TCIA).
In this study, the impact of two factors on generative models is assessed: (i) the size of training data and, (ii) the variability of shapes among individuals. Therefore, two datasets with a different number of shapes and various complexities in the structure are used to assess the model's versatility and performance. A large left ventricle dataset is chosen for its unique structural geometry (inner and outer wall of the left ventricle in a 3D shape), and a smaller liver dataset is used for its wide morphological variations in geometry. In both cases, a random canonical shape was selected as the initial atlas.
\section{Results and Discussion}
This study performs experiments to evaluate the proposed generative shape models, comparing them with each other and with the baseline model RSMP. RSMP is a PCA-based statistical shape model that integrates point cloud-based registration models for shape matching \cite{myronenko2010point}. The evaluation rigorously assesses the generative model's performance across fidelity, diversity, and generalization using established metrics such as specificity, generalization \cite{styner2003evaluation}, distance metrics for dissimilarity, symmetric metrics for inherent symmetry, and statistical metrics for clinical relevance \cite{romero2021clinically}. This comprehensive evaluation ensures a nuanced understanding of the model's performance in diverse aspects.
The evaluation is presented for each component of the model, including shape matching (correspondence establishment) and generation performance.
\begin{figure*}[!t]
    \centering
    \includegraphics[width=1\textwidth]{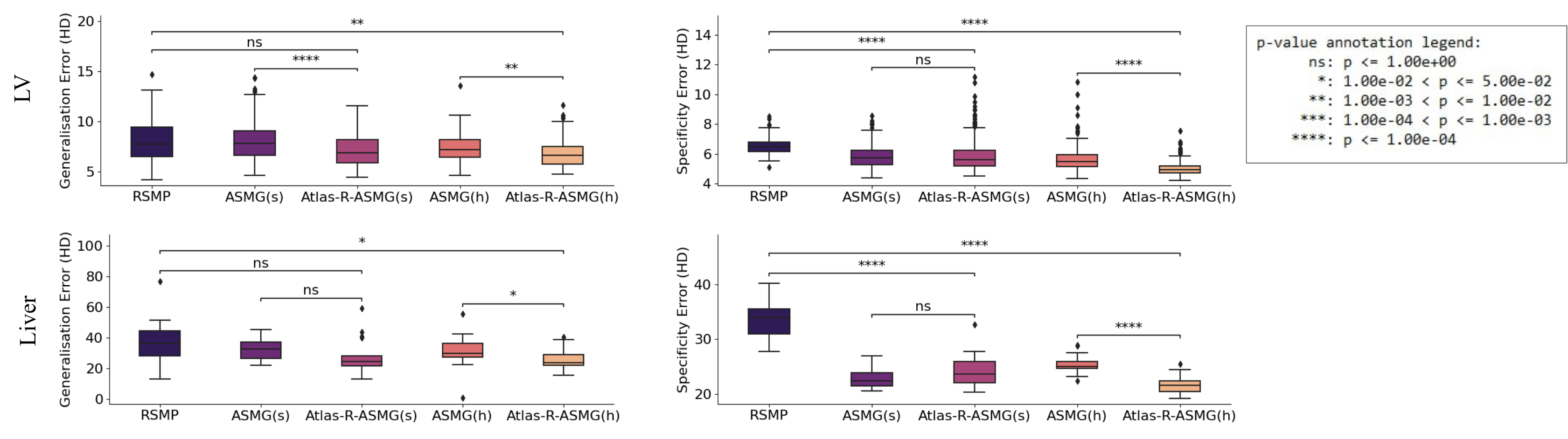}  
    \caption{Generalisation and Specification ability: Boxplots show the generalisation and specificity errors with $HD$ (in $[mm]$) for the different models RSMP, ASM, and Atlas-R-ASMG with different settings s/h, where "s" and "h" refer to sGCN-ATT-VAE and hGCN-ATT-VAE respectively and the models' performance are statistically compared. 
    The top and bottom rows illustrate results on LV and liver datasets, respectively. }
    \label{fig:Gen-Spe}
\end{figure*}
\subsection{Evaluation of Atlas-R-ASMG model}
\subsubsection{Evaluation of Shape Matching}
This section explores the quantitative performance of our Refinable Attention-based Shape Matching (R-ASM) method in two settings (s/hGCN-ATT). We assess its accuracy by comparing it with RSM (Registration-based Shape Matching) \cite{myronenko2010point} and ASM (Attention-based Shape Matching without refinement). This evaluation aims to analyze the significance of surface representation and refinement strategy in attention-based shape matching.
Table \ref{table:Rec-chap5} summarises accuracy on test shapes ($200$ LV cases and $28$ liver cases) in the atlas domain using various methods on LV and liver datasets. Shape quality is assessed with Hausdorff distance ($HD$) and Chamfer distance ($CD$). The results indicate that our learning-based shape matching approach, utilising graph representations, surpasses point-based registration methods. Our method's superiority arises from its spatial-based geometric deep learning and fully-differentiable shape matching, enabling data-driven neighborhood establishment between matched node pairs without requiring any optimisation.
The R-ASM method outperforms the RSM and ASM methods across all the metrics. It means the R-ASM framework presents high-quality shapes in the arbitrary atlas domains, by achieving lower mean $HD$ and $CD$, while the lower standard deviation values indicate its robustness. 
The refinement improves the quality of the normalised shapes over the sGCN-ATT and hGCN-ATT settings in terms of $HD$ and $CD$. 
The lower $HD$/$CD$ metrics of the hGCN-ATT compared to the sGCN-ATT reflect the observation that the sGCN-ATT is still susceptible to producing topological errors, which can be rectified by refinement.

The improvement achieved with refinement over the attention-based shape matching(ASM) is also illustrated in Fig. \ref{fig:scatter-sGCN} and \ref{fig:scatter-hGCN}. The scatter plots show distance metrics $HD$ and $CD$ of ASM and R-ASM models on the LV and liver test cases in two different settings.
In both settings on both LV and liver shapes, the minority of samples have degradation in the metrics, while outliers are generally corrected, particularly for the metric $CD$.
We observe that the R-ASM model has higher performance than ASM across both settings. The improvement is mainly achieved where inputs to the network are not rich in features (i.e. with the sGCN-ATT settings in Fig. \ref{fig:scatter-sGCN}). 


\subsubsection{Evaluation of Generation}

\subsubsection*{Generalisability and Specificity}
A model's generalisability refers to its ability to capture variability in seen data (training) and to generalise that to the unseen data (testing). In generative shape models, this is quantified by the reconstruction error for unseen test data, providing insights into the model's ability to explain unseen shapes and capture the overall variability in shapes. Fig. \ref{fig:Gen-Spe}, in its first column, illustrates our models' generalisation for LV and liver datasets, comparing them with other models. Boxplots showcase generalisation errors in the Hausdorff metric $HD$ (in $[mm]$) for ASMG and Atlas-R-ASMG models in s/hGCN-ATT-VAE settings. 
We observe that Atlas-R-ASMG(h) models significantly outperform RSMP for all the metrics on both datasets, and Atlas-R-ASMG(s) models achieve lower generalisation errors compared to PCA-based model RSMP.
This is due to the limitation of PCA-based models.
These models are prone to overfitting to limited training data, thus not being able to accurately represent anatomies that lie outside of the training distribution, and reconstruction of unseen anatomical structures can subject to significant (large-scale) and subtle (small-scale) variations.
Additionally, generalisation error values associated with Atlas-R-ASMG models are consistently lower when compared with those in ASMG models. The Atlas-R-ASMG(h) model statistically outperformed the RSMG(h) model. 
Hence, learning rich node feature (i.e. hybrid) representations and refinable frameworks is beneficial for developing models with good generalisability. The Atlas-R-ASMG(h) models can capture a greater degree of shape variability for each structure (e.g. LV and liver) than afforded by other models and thereby can synthesise more diverse (in terms of shape) virtual anatomical shape populations than the latter.

Specificity errors are employed to quantify the anatomical plausibility of synthesised virtual shape populations. To calculate these, we measure the distance between each generated sample in the virtual population and the closest (or most similar) shape in the actual population.
In Fig. \ref{fig:Gen-Spe}, the boxplots in the second coulmn illustrate the specificity errors measured with the $HD$ metric for virtual (LV/liver) populations synthesised using different models. 
Results show that Atlas-R-ASMG significantly outperformed RSMP across the two structures.
For both datasets, we observed that Atlas-R-ASMG(h) achieves the highest specificity (i.e. lower specificity errors) and the performance is improved compared to ASMG(h). By refining nodal embeddings in an end-to-end framework, rich latent representations can be learned in graph convolutional networks, and the model can capture more details in the shape-matching and normalisation process. As a result, the VAE generator is then trained with more plausible shapes. Intuitively, this forces a greater degree of plausibility in the shape samples generated from the trained model.

\begin{table*}[!t]
\renewcommand{\arraystretch}{1.2}
\setlength{\tabcolsep}{11pt}
\caption{ Clinical acceptance rates $\mathcal{A}$ [in $\%$] achieved by different generative models for LV and liver volumes. Different settings s/h are considered, where "s" and "h" refer to sGCN-ATT-VAE and hGCN-ATT-VAE, respectively}
\centering
\begin{tabular}{llccccc}
\hline \hline
 Biomarker & Metric & \text{RSMP} & \text{ASMG(s)} & \text{ASMG(h)}  & \text{Atlas-R-ASMG(s)} & \text{Atlas-R-ASMG(h)}  \\ \hline
\multicolumn{1}{l}{\multirow{3}{*}{LV volume}}& $\mathcal{A}_{[\text{min}, \text{max}]}$& $99.75$ & $100 $&   ${99.98} $ & $100$& $100$\\ 

 & $\mathcal{A}_{M \pm 3 B}$ & $86.75$& $98.10$ & ${98.67}$  & $99.12$ & $99.65$ \\ 
 & $\mathcal{A}_{\mu \pm 2 \sigma}$ & $65.85$& $88.65$ & ${88.85}$ & $91.47$ & $95.85$  \\ \hline 
 \multicolumn{1}{l}{\multirow{3}{*}{Liver volume}}& $\mathcal{A}_{[\text{min}, \text{max}]}$& $96.40$ & $100 $&  ${100} $ & $100$ & $99.28$\\ 

 & $\mathcal{A}_{M \pm 3 B}$ & $100$& $100$ & ${100}$  & $100$ & ${100}$  \\ 
 & $\mathcal{A}_{\mu \pm 2 \sigma}$ & $98.56$& $100$ & ${100}$ & $100$ & ${100}$  \\ \hline \hline
\end{tabular}
\label{table:AccepRate-chap5}
\end{table*}
\begin{figure*}
    \centering
    \includegraphics[width=0.9\textwidth]{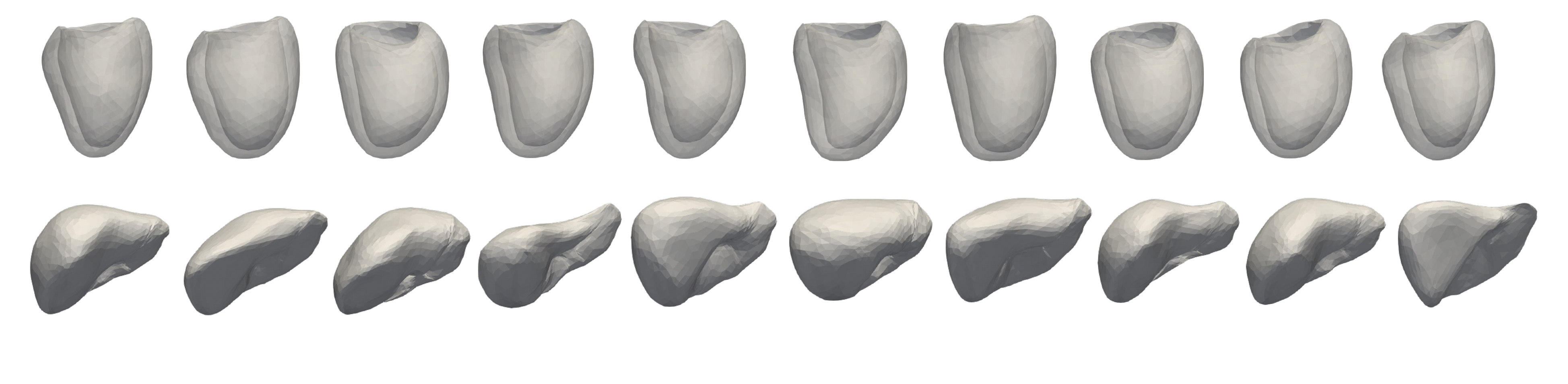}
    \caption{Examples of virtual (LV/liver) samples generated by the Atlas-R-ASMG(h) generator model. }

    \label{fig:synthticshapes}
\end{figure*}
\subsubsection*{Clinical Relevance}
A clinical acceptance rate $\mathcal{A}$ is used as an additional metric to assess virtual cohort anatomical plausibility. It is motivated by the need to preserve clinically relevant anatomical volumetric indices in the synthesised cohorts (as compared to the reference actual population).
Acceptance criteria measure how likely is that synthesised samples in a virtual population with biomarkers (like volume indices) reflect the distribution of the actual biomarker indices. 
Three confidence intervals are used to compute the acceptance rate $\mathcal{A}$. The first criterion involves the range of observed biomarkers in the actual population, denoted as $[\text{min}, \text{max}]$. The second criterion uses Chebyshev's inequality\footnote{A Chebyshev\textquotesingle s inequality is applied when the actual distribution of the different biomarkers is unknown.} for outlier rejection based on the dispersion observed in the original cohort. Assuming unimodality, it establishes the $95\%$ confidence interval\footnote{The same interval contains, at least, $91\%$ of the probability density of the distribution if unimodality cannot be assumed.} $M \pm 3 B$ using the mean ($\mu$), standard deviation ($\sigma$), and mode ($M$) from the actual population, where $B = \sqrt{\sigma^2+ (M-\mu)^2}$ measures the variability across the data.
The third criterion assumes a normal distribution of biomarkers and defines the $95\%$ confidence interval $\mu \pm 2 \sigma$. 

Table \ref{table:AccepRate-chap5} summarises the acceptance rates $\mathcal{A}$ calculated for LV and liver structures in synthesised virtual cohorts. 
The clinical acceptance rates of LV/liver virtual cohorts synthesised by our generative models are higher than the statistical shape RSMP model.
Furthermore, the values estimated for Atlas-R-ASMG are consistently higher than those obtained from ASMG across the LV volume indices, and  there is no huge difference between the values obtained for liver indices. 

In conclusion, the results show that the four models (ASMG(s/h), Atlas-R-ASMG(s/h)) in our generative shape framework better preserve clinical relevance compared to PCA-based models (e.g. RSMP). PCA-based models rely on linear projections, limiting their ability to capture non-linear shape variations. Models trained with PCA exhibit limited generalisation and specificity (as in Fig. \ref{fig:Gen-Spe}), resulting in less anatomically plausible virtual cohorts (as in Table \ref{table:AccepRate-chap5}).

The VAE-based graph convolutional generative shape models (e.g. ASMG and Atlas-R-ASMG), on the other hand, are able to capture non-linear variations in shapes, resulting in virtual cohorts with a greater degree of specificity/anatomical plausibility  (as shown in Fig. \ref{fig:Gen-Spe} and Table \ref{table:AccepRate-chap5}) as well as a better generalisation to unseen shapes (see Fig. \ref{fig:Gen-Spe}). 
Among the VAE-based shape models studied, the Atlas-R-ASMG(h) generator refines nodal embeddings derived from a hybrid representation to optimise training shape alignment on the atlas. This mitigates disorder in the latent space and enhances shape normalisation accuracy, preserving plausibility.
As a result, Atlas-R-ASMG(h) models generate virtual cohorts with improved specificity/anatomical plausibility (see Fig. \ref{fig:Gen-Spe}) and better fidelity in preserving clinically relevant LV/liver index distributions compared to the actual population (as indicated in Table \ref{table:AccepRate-chap5}). Overall, the Atlas-R-ASMG(h) model achieves the highest performance, with lower concurrent specificity and generalization errors for both datasets.

Figure \ref{fig:synthticshapes} demonstrates a visualisation of the generated (synthetic) samples by the Atlas-R-ASMG(h). One can observe the generated meshes demonstrate a level of shape variation that closely resembles that of the actual population for both the left ventricle (LV) and liver shapes. Simultaneously, they retain realistic anatomical shapes at an individual level.



\subsection{Evaluation of mAtlas-R-ASMG model}
In Section \ref{sec:mAtlas}, we discussed how joint-clustering generative models excel in handling shape populations with diverse appearance and anatomical structures, enhancing their generative capabilities.
The liver is a complex organ known for exhibiting significant morphological variations among individuals, making it an ideal candidate for shape clustering and generative analysis. Hence, in this section, we employ the liver dataset to evaluate the joint-clustering generative model and explore the promising approach of multi-atlas construction along with its benefits.
Since earlier results demonstrated significant improvements in anatomical validity with "hybrid representations" of shapes, we employ hybrid representation in the experimental settings for this section.

Table \ref{table:Rec-multiatlas} summarises the accuracy of liver shapes in the shape-matching procedure proposed in the joint-clustering generative model, assessed with $HD$ and $CD$ distance metrics.
The obtained results show that the multiple reference atlases can help improve the accuracy of regressed shapes in different atlas domains. Multiple atlases positively impact the shape-matching process's accuracy, reliability, and robustness by achieving lower mean and standard deviation Hausdorff and Chamfer distances.
By incorporating information from multiple atlases, the model can consider a wider range of anatomical variations, leading to improved matching and generation outcomes.

\begin{table}[!t]
\renewcommand{\arraystretch}{1.2}
\setlength{\tabcolsep}{8pt}

\caption{Shape matching quality in mAtlas-R-ASM model with different number of atlases (i.e. $M$) on liver shapes. Values show the model's performance using two distance metrics $HD$ and $CD$ (mean $\pm$ std) in $[mm]$. \textbf{Bold} values show a significant difference between the methods with a p-value $<0.001$ using the statistical paired t-test.}
\centering
\begin{tabular}{l ccc}
\hline \hline
 & $M=1$ & $M=3$ & $M=5$ \\ \hline
 $HD$& ${20.21 \pm 6.35}$ & ${18.07 \pm 6.39}$  &  $\mathbf{17.33 \pm 5.43}$  \\ \hline
 $CD$ & ${58.14 \pm 21.96}$& ${33.64 \pm 12.33}$ &  $\mathbf{33.84 \pm 10.93}$  \\ \hline \hline
\end{tabular}
\label{table:Rec-multiatlas}
\end{table}
To assess the impact of shape clustering and multi-atlas construction in generative modelling, Table \ref{table:Gen-Spe-multiatlas} presents generalisation and specificity errors of the clustering generative model (i.e. mAtlas-R-ASMG) with different number of clusters. It can be observed that the clustering generative model improves generalisation and specificity metrics (i.e. lower concurrent specificity and generalisation errors). 
This occurs through various mechanisms.
Firstly, it effectively captures complex patterns and relationships within the data by considering multiple clusters. 
This allows the model to more effectively represent the inherent variability and diversity within the dataset, leading to improved generalisation to unseen shapes and enhanced specificity metrics.
The clustering generative model can also incorporate prior knowledge about the class structure (such as initial atlases). By leveraging this prior knowledge, the model can guide the clustering and generative modelling process to capture relevant aspects of the data, characteristics or specific patterns. This integration of prior knowledge enhances the model's generalisation capabilities and improves specificity by identifying class-specific features more accurately.

More importantly, the presented mAtlas-R-ASMG model can be seen as an ensemble of individual clustering and generative models. Each cluster is associated with a specific atlas model and their outputs are combined/weighted to make synthetic shapes. This ensemble approach mitigates errors or biases associated with individual clusters, leading to improved generalisation and specificity by leveraging the collective knowledge of multiple atlas models.


\section{Conclusion}
This paper introduces a novel end-to-end unsupervised generative model aimed at addressing shape matching and generation challenges. 
Our deep learning framework generates virtual anatomical shapes resembling real data from shape surface mesh datasets without prior correspondences.
It handles variable mesh topology in anatomical structures across samples and establishes correspondences from a population of spatially aligned training meshes.
The approach leverages spatial-based graph convolutional neural networks and attention mechanisms to derive a learnable and refinable set of correspondences in the latent space, while concurrently learning an atlas shape. The study demonstrates the model's suitability for synthesizing anatomical shapes through a comparative analysis. Furthermore, extending the model into a joint-clustering generative framework enhances its capabilities in shape analysis and synthesis by accommodating variable topology through multiple atlases, resulting in more robust outcomes.
The proposed generative model constructs a virtual population of single structural shape anatomies. A potential research direction can involve extending the framework as a generative shape compositional framework for complex multi-part structures like biventricular or whole heart anatomies.
\begin{table}[!t]
\renewcommand{\arraystretch}{1.2}
\setlength{\tabcolsep}{8pt}

\caption{Generalisation and Specificity ability in mAtlas-R-ASM model with different number of atlases (i.e. $M$) on liver shapes. Values show the generalisation and specificity errors with $HD$ (mean $\pm$ std) in $[mm]$.}
\centering
\begin{tabular}{l ccc}
\hline \hline
 & $M=1$ & $M=3$ & $M=5$ \\ \hline
 Generalisation& ${25.42 \pm 6.20}$ & ${25.14 \pm 6.06}$  &  ${24.65 \pm 6.00}$  \\ \hline
 Specificity & ${23.24 \pm 4.21}$& ${21.65 \pm 4.09}$ &  ${21.62 \pm 2.80}$  \\ \hline \hline
\end{tabular}
\label{table:Gen-Spe-multiatlas}
\end{table}

\bibliographystyle{IEEEtran}
\bibliography{IEEEabrv,Bibliography}

\end{document}